# ThoughtSource: A central hub for large language model reasoning data

**Simon Ott[1*], Konstantin Hebenstreit[1*], Valentin Liévin[2], Christoffer Egeberg Hother[3], Milad Moradi[1], Maximilian Mayrhauser[1], Robert Praas[1,4], Ole Winther[2], Matthias Samwald[1]**

1) Institute of Artificial Intelligence, Medical University of Vienna, Vienna, Austria

2) Section for Cognitive Systems, Technical University of Denmark, Lyngby, Denmark

3) Department of Clinical Immunology, Copenhagen University Hospital, Copenhagen, Denmark

4) School of Electrical Engineering and Computer Science, The Royal Institute of Technology (KTH), Stockholm, Sweden

* equal contribution

Corresponding author: Matthias Samwald (matthias.samwald [at] meduniwien.ac.at)

## Abstract

Large language models (LLMs) such as GPT-4 have recently demonstrated impressive results across a wide range of tasks. LLMs are still limited, however, in that they frequently fail at complex reasoning, their reasoning processes are opaque, they are prone to 'hallucinate' facts, and there are concerns about their underlying biases. Letting models verbalize reasoning steps as natural language, a technique known as chain-of-thought prompting, has recently been proposed as a way to address some of these issues. Here we present ThoughtSource, a meta-dataset and software library for chain-of-thought (CoT) reasoning. The goal of ThoughtSource is to improve future artificial intelligence systems by facilitating qualitative understanding of CoTs, enabling empirical evaluations, and providing training data. This first release of ThoughtSource integrates seven scientific/medical, three general-domain and five math word question answering datasets.

## Background & summary

The most recent generation of large language models (LLMs) has produced impressive results across a wide range of tasks. Examples of such models include T0 [1], GPT-3 [2], InstructGPT [3] and GPT-4 [4]. These models demonstrated remarkable ability to generate text that is both realistic and coherent, as well as good performance on a broad spectrum of tasks, despite not explicitly being trained on them [3].

However, despite this ability, LLMs are also limited in several ways. They often fail to produce accurate predictions due to their inability to accomplish complex reasoning, such as solving mathematical problems or question answering tasks requiring multi-hop reasoning. Furthermore, they tend to be black boxes, making it difficult to understand how and why predictions are generated. These limitations severely limit the application domains of LLMs and have the potential to cause harm, as lack of explainability and robustness can lead to critical failures and biases when these models are deployed in practice.

One recently proposed method for enabling complex reasoning and generating explanations with LLMs is to force models to explicitly verbalize reasoning steps as natural language, a technique known as chain-of-thought prompting [5,6]. This method improved performance on a variety of tasks and sparked the active development of further refinements [7], such as decomposing problems and structuring reasoning (e.g., least-to-most prompting [8], ReAct [9], self-ask [10], maieutic prompting [11], successive prompting [12]) and/or extending LLM capabilities by leveraging external services for tasks like information retrieval (e.g., self-ask [10], IRCoT [13], DSP [14]). The terminology surrounding these rapidly evolving techniques is not settled, hence in this document, we refer to all approaches that result in a linear sequence of reasoning steps as 'chain-of-thought' (CoT).

Meta-datasets (datasets of datasets) that are easily accessible and standardized have proven useful for training and evaluating versatile LLMs. Examples include SuperGLUE [15] for general-domain language model tasks, BigBIO [16] and BLURB [17] for biomedical tasks, or Pile [18] and ROOTS [19] as text corpora for LLM pre-training. Datasets can be complemented by tools such as PromptSource, which was used to convert a large number of datasets into prompts fit for training and interrogating LLMs. PromptSource facilitated training the highly performant T0 model [1].

Here we present *ThoughtSource*, a meta-dataset and software library for chain-of-thought reasoning in LLMs (https://github.com/OpenBioLink/ThoughtSource). The goals of ThoughtSource are to:

— Facilitate qualitative understanding of CoTs generated by LLMs under various conditions (e.g., across tasks, models and prompts).
— Enable empirical and quantitative evaluation.
— Provide a library of diverse CoT training data for improving performance, robustness, explainability and value-alignment of future LLM-based AI systems.

## Methods

We selected NLP benchmarks for question answering and natural language inference for which pre-existing data for constructing CoTs was available. For some of the datasets, one or multiple additional datasets were used as sources for additional CoTs, allowing for the comparison of different CoT generation methodologies. We created data loader scripts compatible with the Hugging Face datasets library [20] for all datasets. Additionally, we collected metadata of attributes such as descriptions, websites and licenses. We contacted dataset providers and encouraged them to choose an open source/open data license if licensing information was unavailable or unclear.

We implemented two kinds of schemas: 1) source dataset schemas, which are unique to each dataset and provide data close to their original format; and 2) a standardized ThoughtSource schema, which maps all datasets into a common format. The ThoughtSource schema was created by extending the question answering schema of the BigBIO project [16].

We implemented tailored algorithms for converting each dataset because the collected datasets provide explanations in different ways, such as math expressions or structured graph-based explanations. Furthermore, we performed preprocessing such as capitalization and punctuation correction. To recover standard formatted text from pre-tokenized datasets, we reversed the tokenization. This preprocessing was performed only on data in the ThoughtSource schema, while data in the Source schemas was left in their original formatting. All code for running these conversions is available in our Github repository.

We developed a suite of Python libraries and tools for generating novel CoTs and answers by calling LLM APIs, as well as tools for evaluating, comparing and annotating datasets. We built upon the LangChain library (https://github.com/hwchase17/langchain/) for interfacing with a wide variety of external LLM APIs.

This first release of ThoughtSource integrates seven scientific/medical, three general-domain and five math word question answering datasets (Table 1). For every dataset except for PubmedQA and MedQA we provide 'reference CoTs'. We created these reference CoTs by converting rationales provided by original datasets into reasoning chains. These rationales, depending on the dataset, were created by human experts or obtained from crowdsourcing. Furthermore, we added CoTs generated by state-of-the-art LLMs by importing them from previous work, as well as generating them *de-novo* for this work (details below).

*Table 1: Integrated datasets.* *For some core datasets, additional datasets were used as sources for additional CoTs.*

| | **Dataset** | | **License** |
|---|---|---|---|
| | *Scientific and medical question answering* | | |
| WorldTree V2 [21] § | | | AI2 Mercury license |
| EntailmentBank [22] | | | CC BY 4.0 |
| OpenBookQA [23] § | | | Apache License 2.0 |
| MedQA (USMLE) [24] § | Core dataset | | MIT |
| | CoT source: few-shot from Liévin *et al.* [25] | | CC-BY 4.0 |
| | Open ended questions [26] | | MIT |
| MedMCQA [27] § | Core dataset | | MIT |
| | CoT source: few-shot from Liévin *et al.* [25] | | CC-BY 4.0 |
| PubmedQA [28] | Core dataset | | MIT |
| | CoT source: few-shot from Liévin *et al.* [25] | | CC-BY 4.0 |
| MMLU [29] | Core dataset, medical subsets | | MIT |
| | *General-domain question answering* | | |
| CommonsenseQA [30] § | Core dataset | | MIT |
| | CoT source: ECQA [3] | | Community Data License Agreements Sharing license 1.0 |
| | CoT source: few-shot from Wei *et al.* [5], zero-shot from Kojima *et al.* [6] | | Unspecified |
| StrategyQA [31] § | Core dataset | | MIT |
| | CoT source: few-shot from Wei *et al.* [5], zero-shot from Kojima *et al.* [6] | | Unspecified |
| QED [32] | | | CC BY-SA 3.0 |
| | *Math word problems* | | |
| AQUA-RAT [33] | | | Apache 2.0 |
| ASDiv [34] | | | CC BY-NC 4.0 |
| GSM8K [35] | | | MIT |
| MAWPS [36] | | | MIT |
| SVAMP [37] | | | MIT |

§ *for these datasets we generated additional zero-shot CoTs with a variety of LLMs as part of the ThoughtSource-33 subset (license of generated CoTs: MIT)*

Scientific/medical question answering datasets

**WorldTree V2** [21] is one of the most detailed multi-hop science question answering datasets available. Finding the right multiple-choice answers requires a multi-hop inference combining between 1 and 16 facts (average: 6). It contains explanations created by experts in the form of multiple facts. We concatenated these facts and applied a set of rules to improve style and grammaticality to yield reference CoTs that are close to natural language.

**EntailmentBank** [22] contains open-domain science exam questions and answers, along with systematic explanations that show how the correct answer is reached through a series of steps. These steps are organized into a tree structure, known as an entailment tree, which starts with known facts and progresses through intermediate conclusions until the final answer is reached. These entailment trees are also serialized into text-based proofs by traversing the trees. We applied a set of rules to improve style and grammaticality in these proofs to yield reference CoTs that are close to natural language.

**OpenBookQA** [23] contains questions modeled after open-book exams of elementary-level science. They require multi-step reasoning, commonsense knowledge, and a diverse application of core science facts to find the correct answer. The dataset provides over 1,300 core science facts and a mapping to all of the questions. By design, questions in OpenBookQA are answered incorrectly by both retrieval-based and word co-occurrence algorithms. The dataset contains a single-fact explanation of the correct answer for each question, which we adopted to create reference CoTs.

**MedQA** [24] is a free-form multiple-choice OpenQA dataset containing questions from medical board exams in the US (USMLE), Mainland China and Taiwan. We imported the English-language USMLE subset. We have also introduced a version of the dataset wherein the multiple-choice questions have been converted into open-ended questions [26]. Reference CoTs are not provided.

**MedMCQA** [27] is a multiple-choice question answering dataset containing real-world medical entrance exam questions from the All India Institute of Medical Sciences (AIIMS PG) and National Eligibility cum Entrance Test (NEET PG). Answer rationales authored by human experts were integrated as reference CoTs.

**PubmedQA** [28] is a question answering dataset containing biomedical questions extracted from PubMed abstracts that can be answered with yes/no/maybe answers. In addition to the short answer, each question comes with a longer answer, which can be used as reference CoT.

**MMLU** [29] (Massive Multitask Language Understanding) is a compendium of 57 distinct question-and-answer tasks encompassing a wide range of topics. We have selected six subjects particularly related to medical science: anatomy, clinical knowledge, college biology, college medicine, medical genetics, and professional medicine. Reference CoTs are not provided.

## General-domain question answering datasets

**CommonsenseQA** [30] is a collection of multiple-choice questions that test a wide range of general knowledge. We created reference CoTs for the train and validation set derived from the crowd-sourced ECQA dataset[3]. We also added AI-generated reasoning chains generated with few-shot[5] and zero-shot[6] prompting, which are available for the validation split.

**StrategyQA** [31] is a question answering dataset that tests the ability to reason through open-domain questions and provide Yes/No answers. Each example includes a question, a decomposition of the question into reasoning steps, and evidence paragraphs from Wikipedia. The dataset was created through a crowdsourcing process to gather creative and diverse questions. Human-generated freetext reasoning chains are part of the train split of the original dataset and were used as CoTs. The dataset also includes relevant paragraphs from Wikipedia, but these were not included in our CoTs. We extended the StrategyQA dataset with AI-generated CoTs created through few-shot[5] and zero-shot [6] prompting, which are available for the train split.

**QED** [32] is a collection of expert-annotated structured explanations for answers to questions, built upon a subset of the Google Natural Questions dataset. Given a question and a passage from Wikipedia, QED uses linguistic information to represent explanations as a series of interpretable steps, such as referential equality, sentencehood, and entailment. Structured reasoning chains by experts are provided for all examples. To create reference CoTs, we extracted the sentence that entails the answer; statements about referential equality in QED were converted to natural language and added as additional steps in the CoTs (e.g. "The noun phrase [...] in the sentence and the noun phrase [...] in the question refer to the same thing.").

## Math word problem datasets

**Algebra Question Answering with Rationales (AQUA-RAT)** [33] is a large-scale multiple-choice dataset containing algebraic word problems. Each problem consists of a question with five possible answers and a rationale, a step-by-step natural language explanation of the solution. We used natural language explanations as reference CoTs.

**Academia Sinica Diverse (ASDiv) math word problem (MWP) dataset** [34] aims to provide more diverse language patterns and problem types than previous datasets. It covers most of the math topics taught in elementary school. Each MWP is labeled with its grade level (for indicating difficulty), the needed math operation (e.g. division) and includes a short explanation of the solution. ASDiv contains explanations of answers in the form of nested math expressions using common operators such as addition, subtraction, division and multiplication. We generated reference CoTs by converting these math expressions into natural language explanation chains using a rule-based method.

**Grade School Math 8K** (**GSM8K**) [35] contains grade school math word problems. Despite their conceptual simplicity, these problems are more challenging to process than earlier datasets due to their linguistic diversity. The creators of GSM8K instructed crowd workers to write solutions to problems in free text format, which we used as reference CoTs in ThoughtSource, omitting any additional arithmetic specifications.

**Math Word Problems** (**MAWPS**) [36] is an online platform that provides a collection of math word problems. The problems have simple one- or two-line explanations for their solutions. MAWPS includes datasets from various sources, offers tools for automatically creating datasets with specific characteristics as well as the possibility to tune lexical and template overlap. We converted explanatory math expressions to reference CoTs with an approach similar to the one used for ASDiv.

**Simple Variations on Arithmetic Math Word Problems** (**SVAMP**) [37] was created by applying carefully chosen variations to examples from existing datasets, such as ASDiv and MAWPS. These variations make it difficult for language models to solve the problems using simple heuristics, and instead require a deeper understanding and reasoning ability. We converted math expressions to reference CoTs with an approach similar to the one used for ASDiv.

## AI-generated CoTs

**Liévin et al.** CoTs were generated for MedQA, MedMCQA and PubmedQA with the AI systems *text-davinci-002* [3] and *code-davinci-002* [38] (described in detail by co-authors Liévin *et al.* in a separate manuscript [25]).

**Wei et al.** and **Kojima et al.** CoTs for CommonsenseQA and StrategyQA were integrated from previous external studies on few-shot [5] and zero-shot [6] prompting.

**ThoughtSource-33** refers to a collection of 198 items, comprising 33 randomly selected items from each of six datasets: Commonsense QA, MedQA (USMLE), MedMCQA, OpenBookQA, StrategyQA and WorldTree V2. For every item of this collection, we created 60 unique zero-shot CoTs by executing ten different prompting strategies [39] with six models: OpenAI text-davinci-002 [3], OpenAI text-davinci-003 [3], OpenAI GPT-3.5-turbo, OpenAI GPT-4 [4], Flan-T5-XXL [40] and Cohere command-xlarge-nightly (https://docs.cohere.ai/). Since current LLM models are still prone to errors, it should be noted that AI-generated CoTs may contain faulty reasoning.

# Data records

The suggested method for accessing datasets is through programmatic access through our dataloader libraries. A comprehensive guide on how to achieve this is provided on the project's Github repository (https://github.com/OpenBioLink/ThoughtSource). Additionally, a snapshot of the data available through an open license is available on Zenodo [41].

Table 3 shows the example counts, CoT counts and answer types of each dataset. The majority of datasets in the current collection are of the multiple choice answer type. The medical dataset MedMCQA is the largest among all datasets.

*Table 3: Statistics and answer types for all datasets. Note that generated CoTs are not available for all examples, and multiple CoT might have been generated for any given example.*

| Dataset ID | Examples | Examples w. Human Reference CoTs | Examples w. AI-generated CoTs | Number of AI-generated CoTs | Answer type |
|---|---|---|---|---|---|
| AQUA-RAT | 97,975 | 97,975 | 0 | 0 | multiple choice |
| ASDiv | 1218 | 1218 | 0 | 0 | number |
| CommonsenseQA | 12,102 | 10,962 | 1221 | 4417 | multiple choice |
| EntailmentBank | 1840 | 1840 | 0 | 0 | text |
| GSM8K | 8792 | 8792 | 0 | 0 | number |
| MAWPS | 1921 | 1921 | 0 | 0 | number |
| MedQA (USMLE) | 12,723 | 0 | 1273 | 135,640 | multiple choice |
| MedMCQA | 193,155 | 161,558 | 1000 | 106,967 | multiple choice |
| MMLU (medical) | 1242 | 0 | 0 | 0 | multiple choice |
| OpenBookQA | 5957 | 5957 | 100 | 1980 | multiple choice |
| PubmedQA | 1000 | 1000 | 500 | 2500 | multiple choice |
| QED | 6175 | 6175 | 0 | 0 | collection |
| StrategyQA | 2780 | 2290 | 2289 | 6512 | bool |
| SVAMP | 1000 | 1000 | 0 | 0 | number |
| WorldTree V2 | 4367 | 4365 | 100 | 1980 | multiple choice |

Dataset schema

Tables 3–6 provide descriptions and datatypes of the various fields in the ThoughtSource schema. Any performed sample task leads to a generated CoT and answer to the question. Annotations can be added programmatically or through an annotator tool.

*Table 3: Fields of the 'sample' object.*

| Field | Description | Datatype |
|---|---|---|
| id | Unique identifier of object | string |
| ref_id | Identifier of external objects such as documents or other resources | string |
| question | Question of task | string |
| type | Type of the question answering task, currently one of ["multiplechoice", "text", "number", "collection"] | string |
| choices | Set of multiple options containing the gold answer | list(string) |
| context | Additional context for answering the question | string |
| cot | Reference CoT, often human-generated. | list(string) |
| answer | Gold answer of task. Can contain multiple elements if type is collection | list(string) |
| generated_cot | List of generated_cot objects | list(generated_cot_object) |

*Table 4: Fields of the 'generated_cot' object.*

| Field | Description | Datatype |
|---|---|---|
| id | Unique identifier of object | string |
| templates_version | Version of the fragments.json file | string |
| instruction | Identifier of the cot trigger fragment stored in fragments.json | string |
| cot_trigger | Identifier of the cot trigger fragment stored in fragments.json | string |
| cot_trigger_template | Template to specify structure of prompt text | string |
| prompt_text | Full text of prompt used for the CoT generation step | string |
| answers | List of generated answer objects | list(answer_object) |
| cot | Generated chain-of-thought | string |
| author | Name of the author | string |
| date | Date of the chain-of-thought generation | string |
| api_service | Identification of the used api service | string |
| model | Identification of the used language model | string |
| comment | Comment | string |
| annotations | List of annotation objects | list(annotation_object) |

*Table 5: Fields of the 'answer' object.*

| Field | Description | Datatype |
|---|---|---|
| id | Unique identifier of object | string |
| answer_extraction | Identifier of the answer extraction fragment stored in fragments.json | string |
| answer_extraction_template | Template to specify structure of the answer extraction text | string |
| answer_extraction_text | Full text of prompt used for the answer extraction step | string |
| answer | Extracted answer | string |
| correct_answer | True if the extracted answer is equal to the gold answer, else false | bool |

*Table 6: Fields of the 'annotation' object.*

| Field | Description | Datatype |
|---|---|---|
| author | Name of the author | string |
| date | Date of the creation of the annotation | string |
| key | Specifies the label of the annotation | string |
| value | Specifies the value of the annotation | string |

We analyzed the distribution of question and reference CoT field lengths (Fig. 1). MedQA has the longest median question length, while PubMedQA has the longest median CoT length. Several datasets contain outlier CoT with extremely long text lengths. Context fields were only filled for the PubmedQA and QED datasets, with mean context lengths of 116 and 56 tokens, respectively.

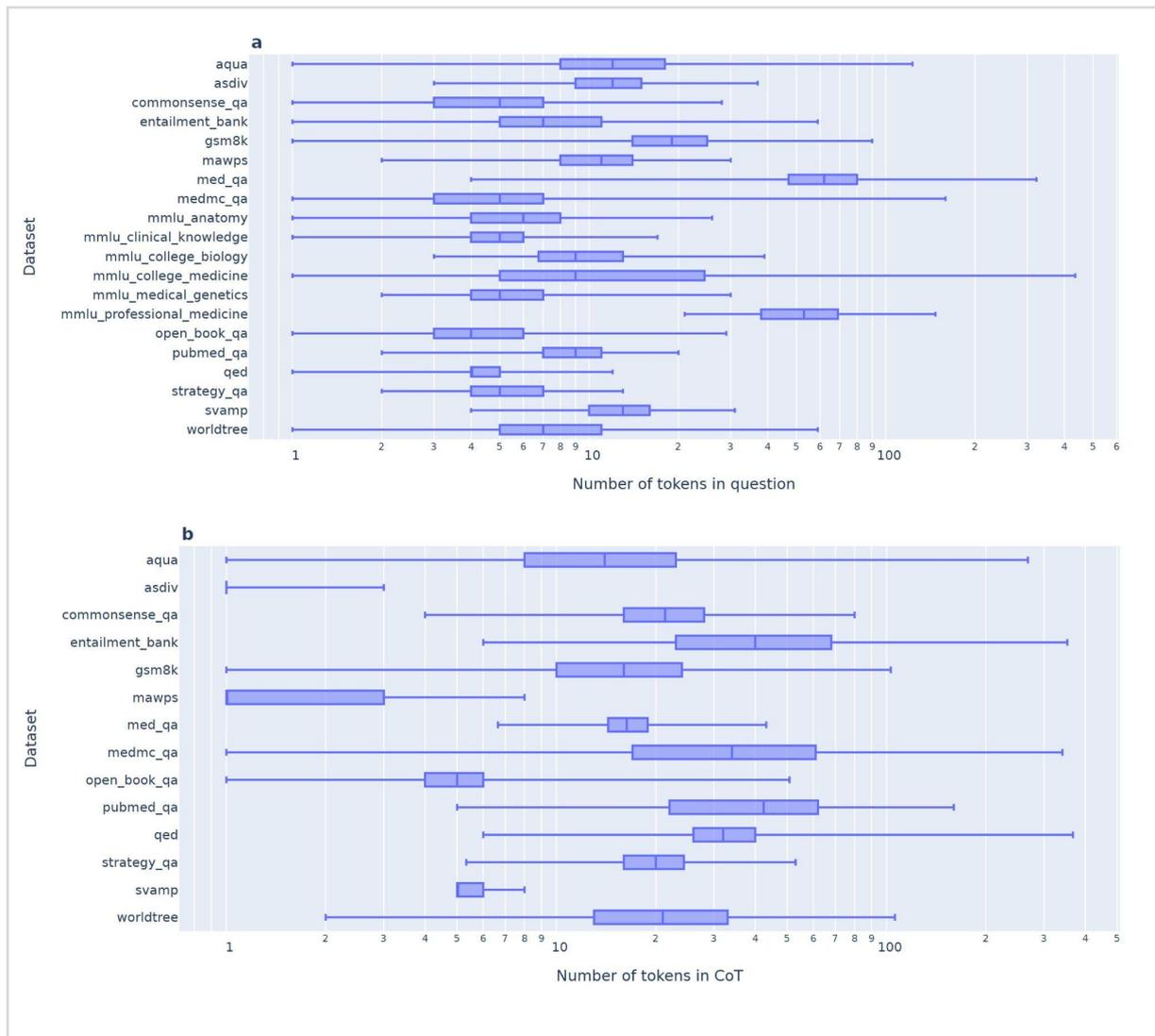

*Figure 1: Distribution of question (a) and reference (b) CoT field lengths.*

## Technical validation

The datasets were reviewed by three team members and issues were tracked on the issue tracker of the associated GitHub repository.

To characterize potential overlaps and relations between datasets, we calculated mutual n-gram overlap using n=3. (Fig. 2) . To quantify the overlap between two sets of n-grams we use the Szymkiewicz–Simpson coefficient (overlap coefficient), which can be interpreted as the proportion of n-grams of the smaller dataset that are contained in the bigger dataset:

$$\text{overlap}(X, Y) = \frac{|X \cap Y|}{\min(|X|, |Y|)}$$

There is an overlap of 1.0 between the set of questions in WorldTree v2 and EntailmentBank. The QA pairs in EntailmentBank were taken from the WorldTree v2 dataset [22], so all the questions in EntailmentBank are a subset of WorldTree v2.

Furthermore, there is significant overlap between the questions contained in ASDiv and SVAMP and those in ASDiv and MAWPS. ASDiv and SVAMP have overlapped questions because a subset of examples from ASDiv was used as seed examples for the creation of SVAMP. For MAWPS and ASDiv, questions were crawled from web resources. The overlap could be due to examples being crawled from the same web resources.

Besides overlaps in questions, we also identified overlaps in reference CoTs. WorldTree v2 provided an initial pool of atomic facts that the annotators could use to construct an explanation tree in EntailmentBank (in addition to creating their own facts). This explains the high overlap of n-grams of CoTs in WorldTree v2 and EntailmentBank. Similarly, a subset of WorldTree v2 facts was used for the creation of explanations in OpenBookQA.

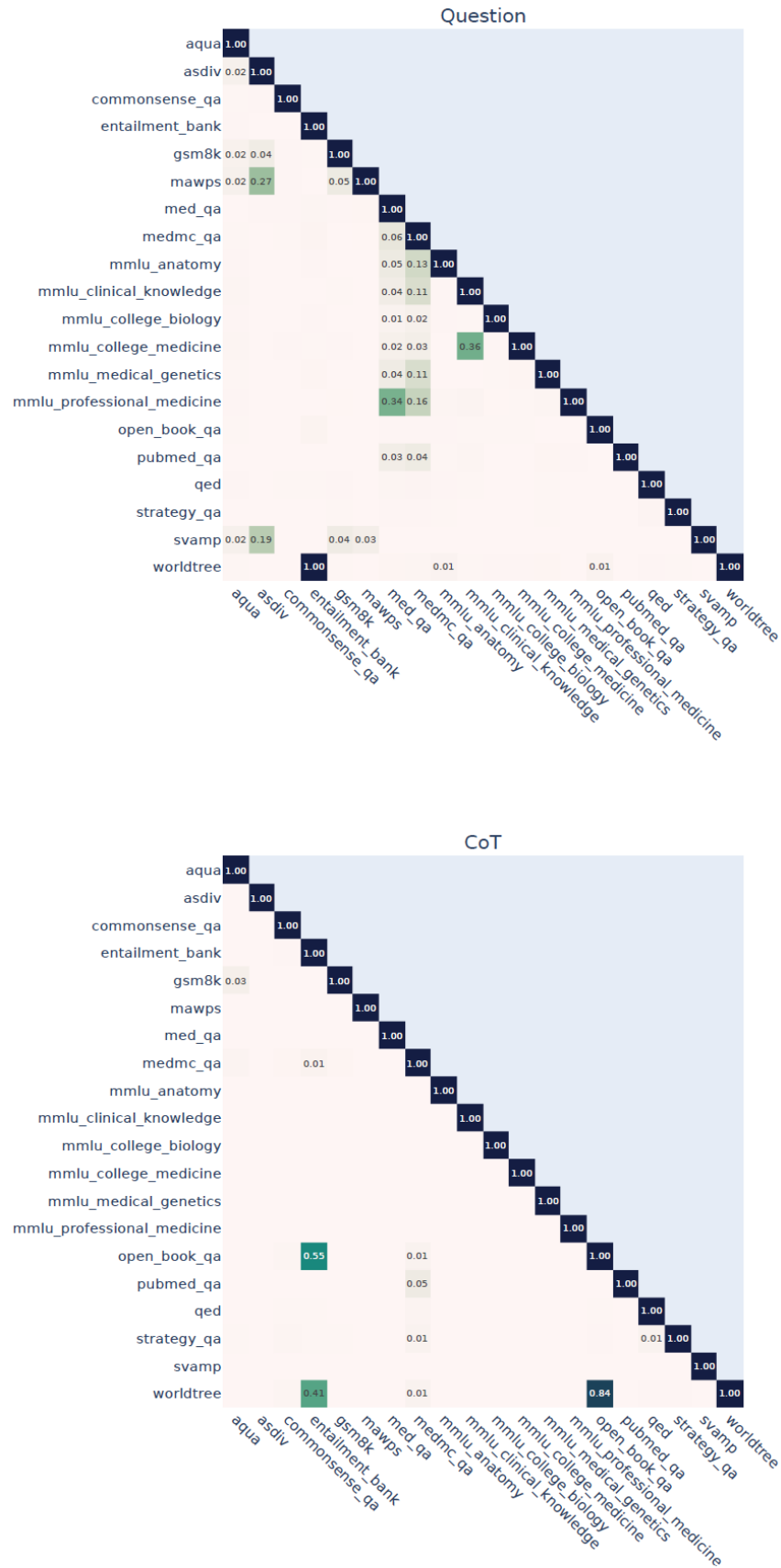

***Figure 2: n-gram overlap in questions and reference CoTs.*** *Overlap is measured by mutual n-gram overlap using n=3, values <0.01 are omitted.*

# Usage notes

Python libraries for accessing and working with data can be downloaded from the Github repository and installed with the *pip* tool. Fig. 3 demonstrates how to load a dataset, randomly sample from the pre-populated data in the dataset, call an external LLM API to generate novel CoTs and answers, automatically evaluate the accuracy of generated answers, and finally save all generated data to a JSON file. Fig. 4 depicts an excerpt of the resulting JSON file.

```python
from cot import Collection

# Load a dataset
collection_worldtree = Collection(["worldtree"])

# Randomly sample 10 rows of train split
collection_worldtree_10 = collection_worldtree.select(split="train", number_samples=10)

# Create a config file for calling OpenAI API to generate new CoTs and answers.
config={
    "instruction_keys": ["qa-01"], # Determines which instructions are used
    "cot_trigger_keys": ["kojima-01"], # Determines which cot triggers are used
    "answer_extraction_keys": ["kojima-A-D"], # Determines which answer extraction
                                    # prompts are used
    "author" : "your_name", # Name of the person responsible for generation
    "api_service": "openai", # Name of the API called ("openai", "huggingface_hub"
                    # or a mock for testing: "mock_api")
    "engine": "text-davinci-002", # Name of the engine used
    "temperature": 0, # Level of randomness in the generated output
    "max_tokens": 512, # Maximum length of output generated by the model
    "api_time_interval": 1.0, # Pause between two api calls in seconds
    "verbose": False, # Determines whether the progress of the generation is printed
    "warn": True, # Determines whether a warnings that external APIs will be called
              # are printed
}

# Generate novel chains-of-thought and answer extractions
collection_worldtree_10.generate(config=config)

# Evaluate accuracy of model predictions
collection_worldtree_10.evaluate()
# Example output: {'accuracy': {'qa-01_kojima-01_kojima-A-D': 0.86}}

# Save all data (including evaluation data) to JSON file
collection_worldtree_10.dump("worldtree_10.json")
```

*Figure 3: **Demonstration of the ThoughtSource API.** Basic functionalities of the data loader, generator and evaluator modules are demonstrated.*

```
{
    "id": "1242",
    "ref_id": "",
    "question": "Which is a characteristic of a sperm cell but not of an egg cell?",
    "type": "multiplechoice",
    "choices": [
        "round shape",
        "presence of a tail",
        "contains genetic information",
        "involved in sexual reproduction"
    ],
    "context": "",
    "cot": [
        "A part of something means a characteristic of something.",
        "A tail is not part of an egg cell.",                          CoT generated by
        "A tail is a part of a sperm cell.",                           human annotators
        "A part of something is present in that something."
    ],
    "answer": [
        "presence of a tail"
    ],
    "generated_cot": [
        {
            "id": "738b54ba-9a20-47e6-b8ff-7cb876103b92",
            "fragments_version": "0.01",
            "instruction": "qa-01",
            "cot_trigger": "kojima-01",
            "cot_trigger_template":"{instruction}\\n\\n{question}\\n{answer_choices}
                                   \\n\\n{cot_trigger}",
            "cot": "We know that both sperm and egg cells are involved in sexual
                   reproduction, so we can eliminate (D). \n\nWe also know that both
 CoT generated    sperm and egg cells contain genetic information, so we can
 by AI model      eliminate (C). \n\nThat leaves us with (A) and (B). \n\nWe know
                   that sperm cells have a tail, but egg cells do not. Therefore, the
                   correct answer is (B).",
            "answers": [
                {
                    "id": "7f7cc26f-a3b3-4b59-9af7-35980514d0c3",
                    "answer_extraction": "kojima-A-D",
                    "answer_extraction_template":
                        "{instruction}\\n\\n{question}\\n{answer_choices}
                        \\n\\n{cot_trigger}{cot}\\n\\n{answer_extraction}",
                    "answer": " B.",
                    "correct_answer": true
                }
            ],
            "author": "your_name",
            "date": "2023/01/12 14:18:57",
            "api_service": "openai",
            "model": "{'name': 'text-davinci-002', 'temperature': 0, 'max_tokens':
                    512}",
            "comment": "",
            "annotation": []
        }
    ]
}
```

*Figure 4: An excerpt of data generated by running the example code. Data for a single question from Worldtree V2 are shown, including human-authored reference CoT, gold-standard answer, an AI-generated CoT and extracted answer, as well as evaluation results. Some fields were omitted for legibility.*

In a zero-shot setup, specific text fragments can be used to prompt question answering and CoT reasoning in LLMs. ThoughtSource includes a curated list of text fragments that can be used to generate novel CoTs (Fig. 5). Where possible, we also mapped individual CoTs in pre-existing CoT datasets to the text fragments that were used in their creation.

```
"instructions": {
    "qa-01": "Answer the following question through step-by-step reasoning.",
    "qa-02": "Answer the following question through careful, concise step-by-step
              reasoning.",
    "qa-03": "Answer the following question through careful, concise step-by-step
              reasoning. Avoid making up wrong statements. If the question does not
              make sense or cannot be answered, write \"I cannot answer the
              question\".
              If you do not have a good answer, write \"I do not have a good answer\".
              If you are uncertain, write \"I am uncertain about this\".",
    [...]
},
"cot_triggers": {
    "kojima-01": "Answer: Let's think step by step.",
    "kojima-02": "Answer: We should think about this step by step.",
    "kojima-03": "Answer: First,",
    "kojima-04": "Answer: Before we dive into the answer,",
    [...]
    "lievin-01": "Answer: Let's derive the differential diagnosis step by step.",
    "lievin-02": "Answer: Let's use step by step inductive reasoning, given the
                  medical nature of the question.",
    [...]
    "lievin-26": "Answer: Let's follow a Bayesian step by step approach.",
    "lievin-27": "Answer: Let's reflect on each option from the least likely to the
                  most likely.",
    "lievin-28": "Answer: Let's use step by step Bayesian reasoning, given the
                  medical nature of the question."
},
"answer_extractions":{
    "kojima-01": "Therefore, the answer is",
    "kojima-02": "Therefore,",
    "kojima-03": "The answer is",
    "kojima-numerals": "Therefore, the answer (arabic numerals) is",
    "kojima-yes-no": "Therefore, the answer (Yes or No) is",
    "kojima-A-C": "Therefore, among A through C, the answer is",
    "kojima-A-D": "Therefore, among A through D, the answer is",
    [...]
}
```

*Figure 5: An excerpt of the collection of prompt fragments. These fragments can be used to build prompts for interacting with LLMs, allowing for empirical testing of how different prompts affect model performance.*

We provide two web-based interfaces for exploring and annotating ThoughtSource data, the *Dataset Viewer* and the *Annotator*. The Dataset Viewer is a simple interface for exploring dataset contents. The Annotator (Fig. 6) allows you to upload specific subsets of a dataset, provides convenience functions for highlighting similarities between different generated CoTs and the correctness of generated answers, and allows you to annotate individual CoTs interactively. The annotator facilitates identifying strengths and weaknesses of different CoTs. Annotations can be used for downstream model evaluation and further improving the capabilities of AI models through fine-tuning / reinforcement learning.

[Screenshot of ThoughtSource Annotator web interface showing Question 1242 (test): A 28-year-old woman, gravida 1, para 0, at 20 weeks' gestation comes to the physician with her husband for a prenatal visit. Her pregnancy has been uncomplicated. They are planning to travel to Ethiopia next month to visit the husband's family. Medications include folic acid and an iron supplement. Vital signs are within the normal range. Abdominal examination shows a uterus that is consistent with a 20-week gestation. Which of the following drugs is most suitable for pre-exposure prophylaxis against malaria? Options A-E with three reasoning chains displayed side by side, with highlighting showing similar text between them, and annotation checkboxes for Incorrect reasoning, Insufficient knowledge, Incorrect reading comprehension, Too verbose.]

*Figure 6: The ThoughtSource Annotator.* The web-based interface allows for convenient inspection and annotation of reasoning chains and answers. Text that is similar between CoTs can be automatically highlighted based on an easily adjustable similarity threshold, facilitating a better understanding of similarities and differences of different reasoning chains.

All tools and libraries, as well as more detailed demonstration notebooks, can be found on the project Github page.

We plan to add more datasets and generated CoTs to the ThoughtSource repository, and we welcome outside contributions. Novel CoTs for existing core datasets can be generated and shared through the ThoughtSource APIs and JSON files. Completely new datasets can also be added, as described in the Github repository's contribution guide.

## Code availability

All code, data and tools are openly available at https://github.com/OpenBioLink/ThoughtSource, a snapshot is archived on Zenodo at https://doi.org/10.5281/zenodo.8155593 [41]. Our code and data are licensed under an MIT license, while data adapted from existing datasets are available under the licenses of their respective sources.

## Acknowledgements

We thank primary dataset contributors who assisted with assembling the ThoughtSource meta-dataset.

## Author contributions

S.O. and K.H. wrote the code for accessing, converting, generating and analyzing datasets, and wrote parts of the manuscript and documentation.

V.L., C.E. and O.W. generated and analyzed CoT data for medical datasets.

M.Ma. wrote the code of the annotator software.

M.Mo. wrote a first prototype of code for accessing and converting datasets.

R.P. contributed to improving code and documentation quality.

M.S. conceived and supervised the project and wrote parts of the manuscript and documentation.

All authors have read and approved the final manuscript.

## Competing interests

The authors declare that there are no conflicts of interest.